\documentclass[letterpaper]{article} 
\usepackage{aaai25}  
\usepackage{times}  
\usepackage{helvet}  
\usepackage{courier}  
\usepackage[hyphens]{url}  
\usepackage{graphicx} 
\usepackage{natbib}  
\usepackage{caption} 
\usepackage{algorithm}
\usepackage{algorithmic}

\usepackage{newfloat}
\usepackage{listings}
\DeclareCaptionStyle{ruled}{labelfont=normalfont,labelsep=colon,strut=off} 
\floatstyle{ruled}
\newfloat{listing}{tb}{lst}{}
\floatname{listing}{Listing}
\usepackage{amsmath}
\usepackage{booktabs}
\usepackage{multirow}

\nocopyright 
\title{OMEGAS: Object Mesh Extraction from Large Scenes Guided by Gaussian Segmentation}
\author{
    Lizhi Wang\equalcontrib,
    Feng Zhou\equalcontrib,
    Bo Yu,
    Pu Cao,
    Jianqin Yin\thanks{Corresponding author},
}
\affiliations{
    Beijing University of Posts and Telecommunications\\

    Beijing, China\\
    wanglizhi@bupt.edu.cn
}

\begin{document}
\maketitle

\begin{abstract}
Recent advancements in 3D reconstruction technologies have paved the way for high-quality and real-time rendering of complex 3D scenes. Despite these achievements, a notable challenge persists: it is difficult to precisely reconstruct specific objects from large scenes. Current scene reconstruction techniques frequently result in the loss of object detail textures and are unable to reconstruct object portions that are occluded or unseen in views.
To address this challenge, we delve into the meticulous 3D reconstruction of specific objects within large scenes and propose a framework termed \textbf{OMEGAS}: \textbf{O}bject \textbf{M}esh \textbf{E}xtraction from Large Scenes Guided by \textbf{GA}ussian \textbf{S}egmentation. Specifically, we proposed a novel 3D target segmentation technique based on 2D Gaussian Splatting, which segments 3D consistent target masks in multi-view scene images and generates a preliminary target model. Moreover, to reconstruct the unseen portions of the target, we propose a novel target replenishment technique driven by large-scale generative diffusion priors. We demonstrate that our method can accurately reconstruct specific targets from large scenes, both quantitatively and qualitatively. Our experiments show that OMEGAS significantly outperforms existing reconstruction methods across various scenarios.
\end{abstract}

\section{Introduction}

In recent years, the field of 3D reconstruction has emerged as a pivotal area of research, driven by its profound applications across a diverse range of disciplines, including robotics~\cite{siciliano2008springer}, architectural design~\cite{caetano2020computational}, virtual reality~\cite{xiong2021augmented}, and so on. The community has successfully achieved high-quality, real-time reconstruction and rendering of complex 3D scenes, largely due to the advancement of 3D rendering-based models~\cite{mildenhall2020nerf, barron2022mip, yu2021pixelnerf, kerbl20233d, wu20234d, yang2023deformable}.
\begin{figure}[h]
  \centering
  \includegraphics[width=\columnwidth]{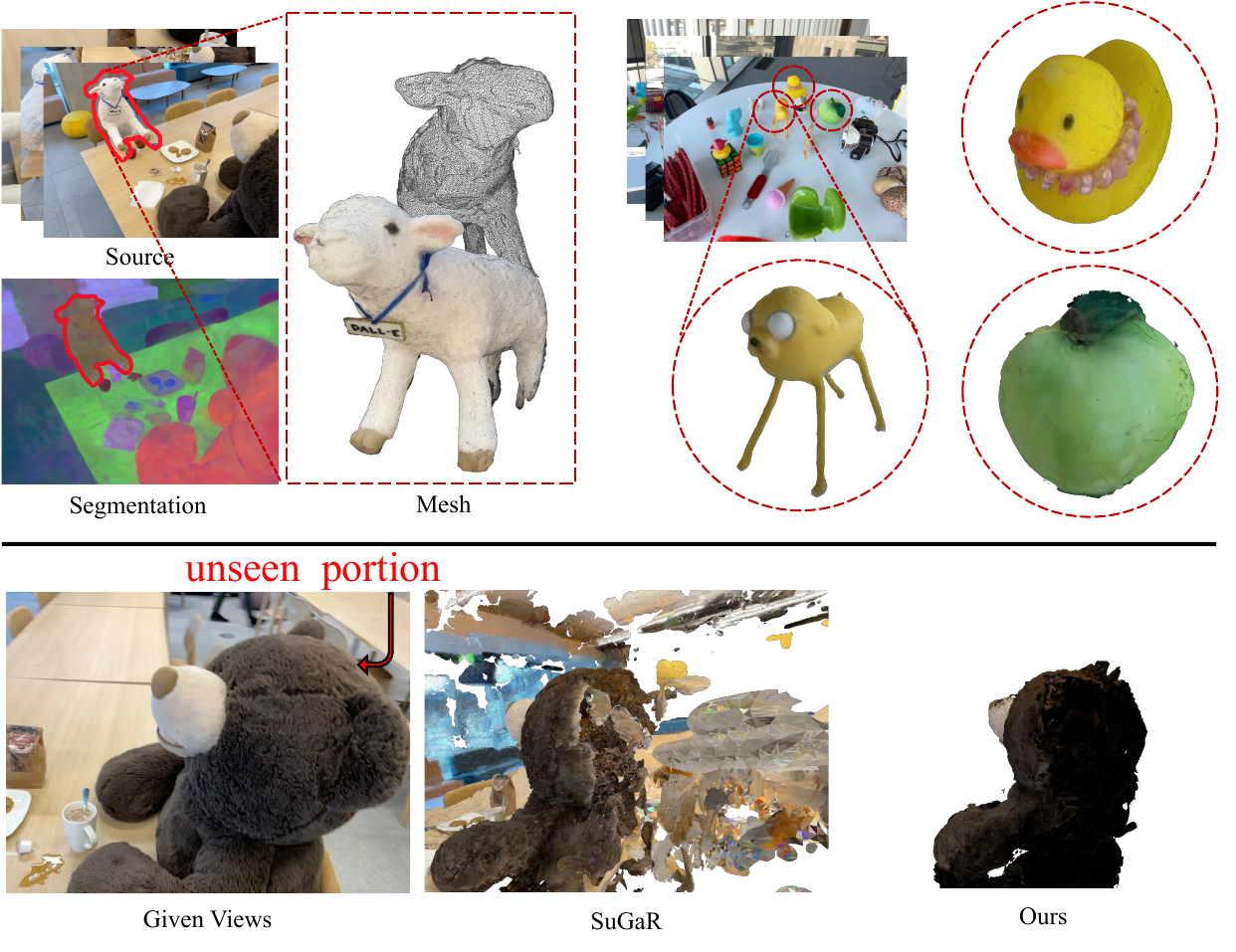}
  \caption{OMEGAS segments and generates high-quality meshes for specified objects in open-world scenes (top). OMEGAS can effectively reconstruct the unseen portion of the target (bottom).}
  \label{fig:intro_1}
\end{figure}

However, existing methods struggle to reconstruct a specific target object's 3D mesh in high-quality given scene images, which is somehow more solid and straightforward in downstream applications like virtual game modeling. This challenge is particularly evident in two key aspects.
First, the reconstruction of entire scenes often leads to a compromise in the quality of specific object reconstruction. Besides, certain parts of a specific object in a scene are frequently occluded or invisible from any perspective, making their reconstruction challenging with current methods, as illustrated in Figure~\ref{fig:intro_1} bottom.

In this paper, we propose a novel and effective framework that reconstructs the 3D mesh of the target object given multi-view open-world scene images, termed OMEGAS: Object Mesh Extraction from Large Scenes Guided by GAussian Segmentation. As shown in Figure~\ref{fig:intro_1} top, given the multi-view scene images, our framework can freely select and segment the target object from images and extract the complete 3D object mesh. 

Specifically, we propose a novel 3D target segmentation technique based on 2D Gaussian Splatting (2DGS)~\cite{huang20242d}. Inspired by Gaussian Grouping~\cite{Ye2023GaussianGS}, we first utilize the SAM~\cite{kirillov2023segment} to provide initial target segmentation masks in multi-view scenes. Then, we input the masks along with the scene images into a 2DGS model and introduce a unique compact identity vector as a supervisory signal, enabling 3D segmentation within the 2DGS space. By leveraging the 3D consistency of the 2DGS model, we can iteratively optimize the segmentation results of the target within the scene, ultimately obtaining precise 2D segmentation results along with an initial 3D model of the target.

Subsequently, to address the occluded portions of the target that cannot be directly reconstructed, we leverage large-scale, open-world generative priors such as Stable Diffusion~\cite{rombach2022high}. Specifically, we segment the target from the original perspective and then use these target-only images to personalize Stable Diffusion. This approach allows for the concentration of knowledge on the specific target. Our goal is to utilize the customized Stable Diffusion to assist inpainting the incomplete faces of the target Gaussian model. In this process, we propose a novel mask generation technique based on the diffusion denoising process to mark the under-constructed portions of the novel-view rendered images. Ultimately, we further train the 2DGS model with additional inpainted view images to obtain the  final, fully completed target mesh.

Extensive experiments demonstrate the superior performance of our framework, capable of reconstructing the target object mesh with high precision. 

In summary, our contributions are:
\begin{itemize}
\item We propose OMEGAS, an effective framework to extract meshes of specified objects through multi-view 2D images in open-world scenes. 

\item We propose a novel 3D target segmentation technique based on 2D Gaussian Splatting to extract the target 3D model from multi-view scene images and create 3D-consistent target masks. Furthermore, we introduce a novel target replenishment technique by leveraging large-scale generative diffusion priors to adaptively optimize the unseen portions of the target.

\item Extensive experiments reveal that OMEGAS operates effectively in a range of open-world scenes and significantly surpasses the existing approach in performance. Specifically, our approach shows superiority in both texture details and occlusion robustness on target object reconstruction.
\end{itemize}

\section{Related Works}
\subsection{Segmentation of Rendering-based 3D Models}
Due to the intrinsic implicit nature of the rendering-based model, it's hard to do normal data-driven training for semantic segmentation like explicit 3D models (\emph{e.g.}, 3D point-cloud segmentation~\cite{xu2020squeezesegv3, fan2021scf} ). To address this issue, Spin-NeRF~\cite{fan2021scf} first proposes a method that segments specific objects in NeRF and achieves inpainting in different views with 3D consistency. Recently, 3D Gaussian Splatting established its effectiveness in the reconstruction task exhibiting high inference speeds and remarkable quality. Following Spin-NeRF, Gaussian Grouping~\cite{Ye2023GaussianGS} extends the object-oriented concept from Spin-NeRF to 3D Gaussian Splatting, enabling the joint reconstruction and segmentation of anything in open-world 3D scenes and various 3D editing tasks.

\begin{figure*}[h]
  \centering
  \includegraphics[width=\linewidth]{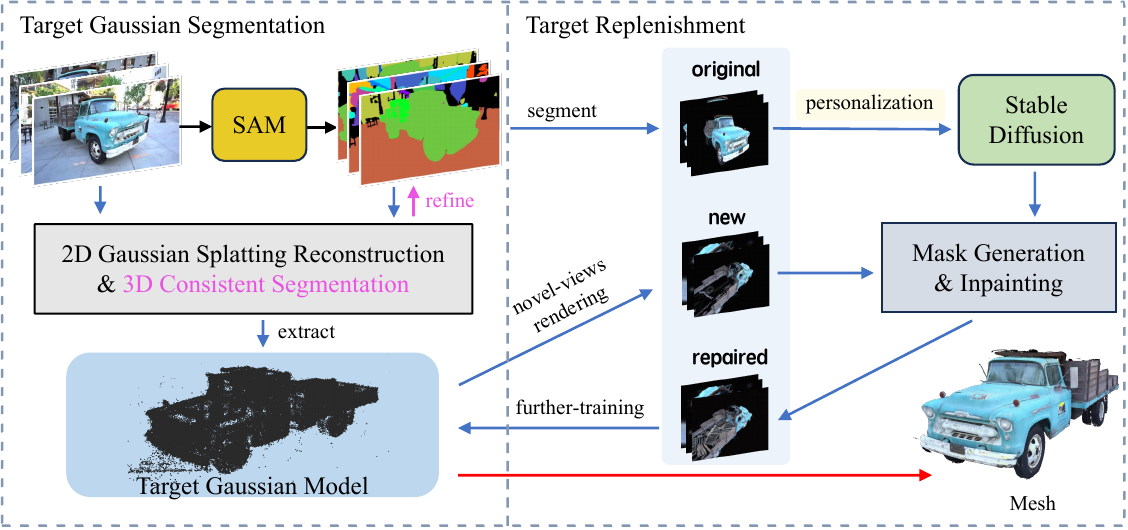}
  \caption{The framework of \textbf{OMEGAS}. OMEGAS comprises two stages: \textbf{Target Gaussian Segmentation} to segment target 2DGS model from multi-view images and provide accurate 3D consistent masks; \textbf{Target Replenishment} to optimize the target model by personalized Stable Diffusion with a Mask-generation \& Inpainting process. The final object mesh is extracted from the optimized target Gaussian model.}

  \label{pipeline}
\end{figure*}

\subsection{Mesh Extraction from Images}
In the early days, the development of Structure-from-motion (SfM)~\cite{snavely2006photo} and Multi-View Stereo (MVS)~\cite{goesele2007multi} allows for 3D reconstruction from multi-view images. More recently, owing to the development of neural network-based 3D reconstruction method~\cite{mildenhall2020nerf, kerbl20233d}, various approaches have explored integrating rendering-based models with mesh reconstruction~\cite{yang2022neumesh, darmon2022improving, li2023neuralangelo, oechsle2021unisurf, wang2021neus}. For example, some works optimize neural signed distance functions (SDF) by training neural radiance fields (NeRF) in which the density is derived as a differentiable transformation of the SDF~\cite{li2023neuralangelo, oechsle2021unisurf}. A triangle mesh can finally be reconstructed from the SDF by applying the Marching Cubes algorithm~\cite{Lorensen1987MarchingCA}. Notably, SuGaR~\cite{Guedon2023SuGaRSG} introduced the pioneering method for high-precision scene mesh reconstruction from a 3DGS model. In contrast, our frameworks focus on specific target mesh reconstruction from scenes.

\subsection{3D Reconstruction from Partial Views}
Vanilla methods like NeRF~\cite{mildenhall2020nerf} or 3DGS~\cite{Ye2023GaussianGS} struggle in reconstruction under partial-view settings. Some research efforts focused on incorporating auxiliary priors, such as depth maps~\cite{song2023darf, wang2023sparsenerf}, information theory~\cite{kim2022infonerf}, symmetry~\cite{seo2023flipnerf}, and continuity~\cite{niemeyer2022regnerf} to infer missing information. However, these approaches tend to be either too simplistic or overly specialized for certain scenes, resulting in poor generalization. 

The recent progress in text-to-image diffusion models~\cite{rombach2022high} and their tailored applications~\cite{poole2022dreamfusion, wang2024prolificdreamer} in the 3D field make it possible for reasonable novel-view synthesis, paving the way for impactful applications such as single-view 3D asserts generation (image-to-3d)~\cite{Liu2023Zero1to3ZO, liu2024one, tang2023dreamgaussian} and sparse reconstruction~\cite{wynn2023diffusionerf, xie2023sparsefusion, liu2023deceptive, wu2024reconfusion, yang2024gaussianobject}. However, existing sparse reconstruction methods are designed for scenarios where camera views are sparsely distributed across a 360 range. In contrast, our approach aims to reconstruct partially occluded targets by focusing on cases where views are densely clustered within the visible range of the target, while the occluded views remain entirely absent.


\section{Method}

Our framework aims to reconstruct the precise mesh of the target object from multi-view scene images. It comprises two main components: Target Gaussian Segmentation (TGS) to segment the 3D target from multi-view scene images and Target Replenishment (TR) to optimize the target by large-scale diffusion priors, as illustrated in Figure~\ref{pipeline}. 

The rest of this section is organized as follows: in section \emph{Preliminaries}, we introduce some basic concepts involved in this section, including 2D Gaussian Splatting~\cite{huang20242d} and the generative priors. In section \emph{Target Gaussian Segmention}, we introduce the technique that segments the initial 3D target from multi-view images by 2D Gaussian Splatting. In section \emph{Target Replenishment}, we introduce the process that adaptively optimizes the target model by diffusion priors. 

\subsection{Preliminaries} 
\label{sec:3.1}
\textbf{2D Gaussian Splatting (2DGS)~\cite{huang20242d}. } 
Due to the multi-view inconsistent nature of 3D Gaussians, 3D Gaussian Splatting (3DGS) fails to accurately represent surfaces. Thus,~\citeauthor{huang20242d} proposed 2D Gaussian Splatting, which collapses the 3D volume into a set of 2D oriented planar Gaussian disks and provides view-consistent geometry while modeling surfaces intrinsically. Specifically, a 2D Gaussian is defined in a local tangent plane in world space, which is parameterized:
\begin{equation}
P(u, v) = \mathbf{p}_k + s_u \mathbf{t}_u u + s_v \mathbf{t}_v v = \mathbf{H} \begin{pmatrix} u , v , 1, 1 \end{pmatrix}^{\top}
\end{equation}
\begin{equation}
\text{where } \mathbf{H} = \begin{bmatrix}
    s_u \mathbf{t}_u & s_v \mathbf{t}_v & 0 & \mathbf{p}_k \\
    0 & 0 & 0 & 1 
\end{bmatrix} 
= \begin{bmatrix}
    \mathbf{RS} & \mathbf{p}_k \\
    0 & 1 
\end{bmatrix}
\end{equation}
where a 2D Gaussian plane $P(u, v)$ is defined by its central point $
\mathbf{p}_k$, principal tangential vectors $t_u$ and $t_v$, scaling vectors $s_u$ and $s_v$. $\mathbf{H} \in 4 \times 4$ is the homogeneous transformation matrix representing the geometry of the 2D Gaussian.  For the point $\mathbf{u}=(u,v)$ in $uv$ space, its 2D Gaussian value is evaluated by standard Gaussian:
\begin{equation}
\mathbf{G}(\mathbf{u}) = \exp\left(-\frac{u^2+v^2}{2}\right)
    \label{eq:gaussian-2d}
\end{equation}
The center $\mathbf{p}_k$, scaling $(s_u,s_v)$, and the rotation  $(\mathbf{t}_u, \mathbf{t}_v)$ are learnable parameters.
Our framework harnesses the 2DGS model as the target reconstruction carrier due to its meticulous object reconstruction capabilities.

\noindent\textbf{Text-to-Image Generative Priors \& Personalization. } In the past few years, with the advent of diffusion-based generative techniques, the community has developed numerous mature open-world large-scale generative models, \emph{e.g.,} Stable Diffusion~\cite{rombach2022high}. Technically, given input image $x\in\mathbf{R}^{H\times W\times 3 }$, an encoder of variational auto-encoder (VAE) $\mathcal{E}$ converts it into a latent representation $z_0\in\mathbf{R}^{h\times w\times c}$, where image is downsampled by a factor $f$ and channels are increased to $c$. Then a text-conditioned UNet $\epsilon_\theta$ in latent space is used to predict the noise of each timestamp $t$ from $z_t$ and prompt $y$ to recover $z_0$, where $\theta$ is the param of UNet. After the denoising process, VAE's decoder $\mathcal{D}$ transforms $z_0$ to image space to generate the image.

The personalization of diffusion priors~\cite{ruiz2023dreambooth, cao2024controllable} is a technique that concentrates prior knowledge on a specific object by transforming or fine-tuning UNet parameters $\theta$ into a form specific to the given few object images, represented as $\epsilon_{\theta^*}$.

\subsection{Target Gaussian Segmentation (TGS)}
As depicted in the left part of Figure~\ref{pipeline}, we begin by introducing a novel technique known as Target Gaussian Segmentation. This method segments 3D-consistent target masks from multi-view images and extracts a preliminary 2DGS model of the target. The process is as follows.

\begin{figure}[t]
  \centering
  \includegraphics[width=\columnwidth]{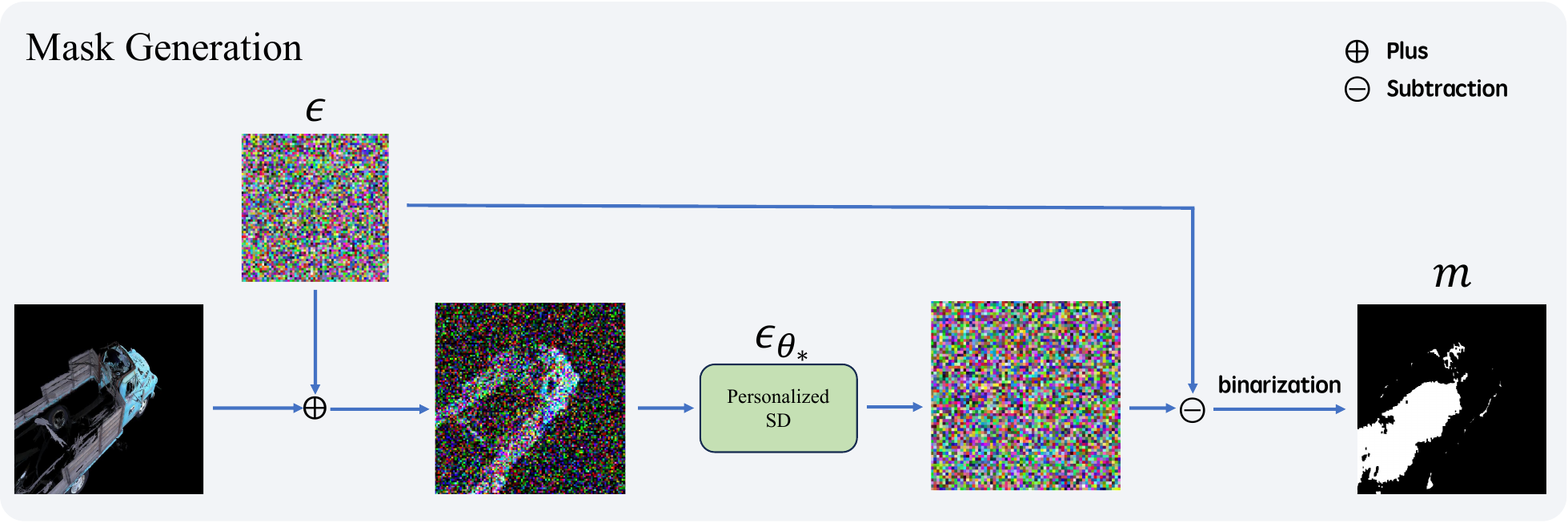}
  \caption{Mask generation process. }
  \label{fig:method_2}
\end{figure}

\noindent\textbf{Scene Segmentation by SAM.}
We first conduct preliminary target consistency segmentation on scene images from different views based on SAM's capabilities in open-world segmentation for intricate scenes. Specifically, following the approach in~\cite{Ye2023GaussianGS}, we treat the scene images from divergent views as a continuous video sequence and input them into SAM for initial segmentation. Next, by applying a zero-shot tracker~\cite{cheng2023tracking} to these segmentation results, we obtain unique object IDs ranging from 0 to 255. The segmented outcomes are preserved as gray-scale images, with each object's ID represented in corresponding gray-scale values.

\noindent\textbf{Target 2DGS Reconstruction \& 3D Consistent Segmentation. } 
We leverage the gray-scale images to guide the segmentation of 2DGS. We treat the gray scale as attributes similar to RGB colors for training 2D Gaussians. Similar to representing Gaussian colors using spherical harmonics coefficients, we add identity vectors to 2D Gaussians following~\cite{mirzaei2023spin, Ye2023GaussianGS}. To reduce the memory usage and training time, we set the identity vectors as length 8, encoding segmentation labels ranging from 0 to 255. 
When a 2D Gaussian is observed from a slanted views, the object-space low-pass filter is utilized:
\begin{equation}
    \hat{\mathbf{G}}(\mathbf{x}) =\max\left\{ \mathbf{G}(\mathbf{u}(\mathbf{x})),  \mathbf{G}(\frac{\mathbf{x}-\mathbf{c}}{\sigma})\right\}
\end{equation}
where $\mathbf{c}$ is the projection of center $\mathbf{p}_k$. 

By conducting differentiable rendering of identity vectors, similar to rendering colors, by blending $\mathcal{N}$ ordered Gaussians on overlapping pixels, we can calculate the identity vectors $O$ of pixels:
\begin{equation}
\label{eq:2d_id}
O(\mathbf{x}) = \sum_{i \in \mathcal{N}} o_i \alpha'_i \hat{\mathbf{G}}_i(\mathbf{u}(\mathbf{x})) \prod_{j=1}^{i-1} (1 - \alpha'_j\hat{\mathbf{G}}_j(\mathbf{u}(\mathbf{x})))
\end{equation}where \(o_i\) represents the identity vector of each Gaussian, and $\alpha'_i$ is given by evaluating the opacity of 2D Gaussians multiplied by the opacity of each point. Unlike colors, the identity vectors of the same object does not change with different viewpoints, so we set the SH degree to 0, reducing computational complexity.

\begin{figure*}[h]
  \includegraphics[width=\linewidth]{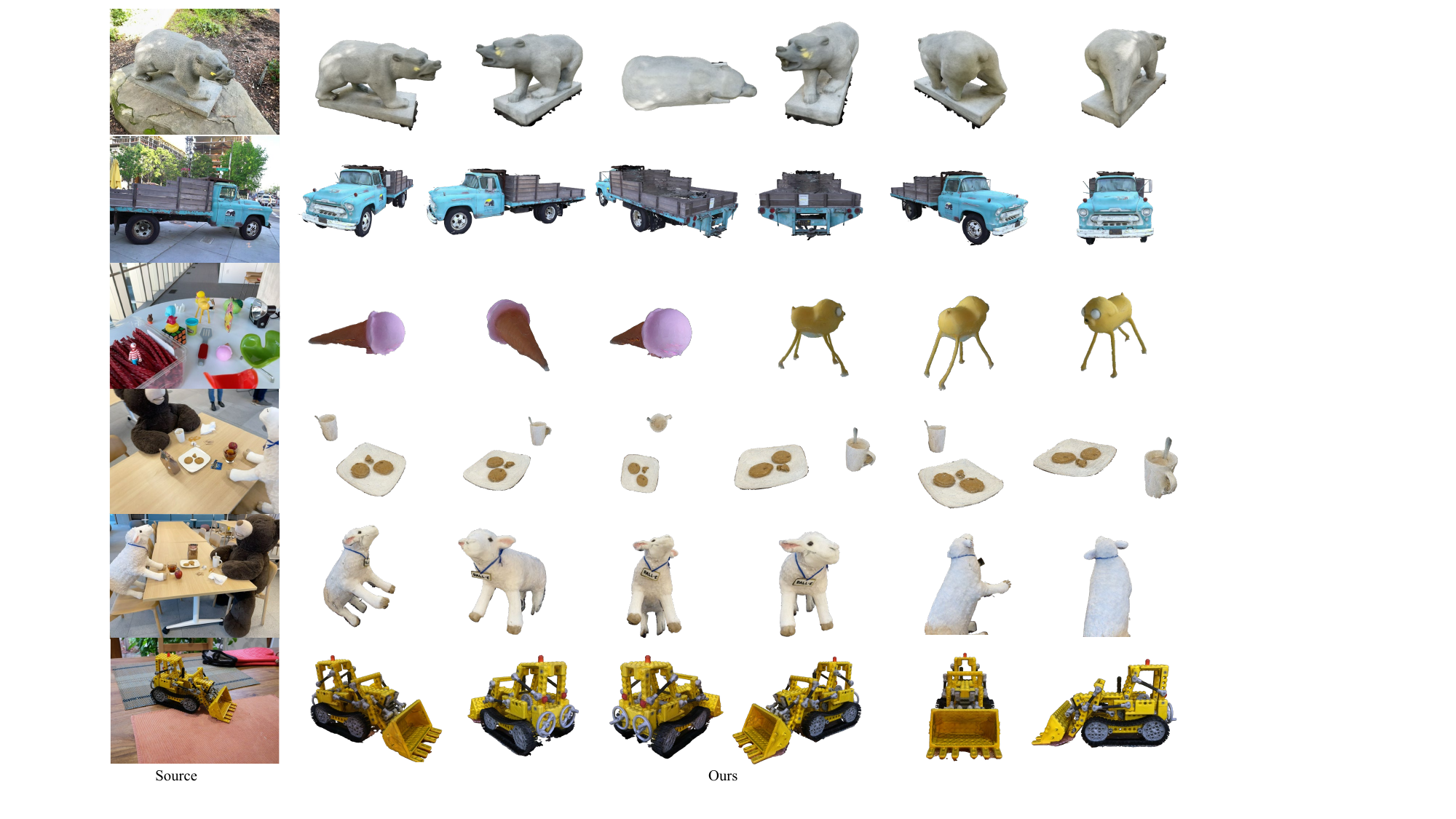}
  \caption{Results of meshes with OMEGAS on the Instruct-NeRF2NeRF dataset(line 1), Tanks\&Temples dataset(line 2), LERF dataset(line 3,4,5) and Mip-Nerf360 dataset(line 6). }
  \label{fig:exp_2}
\end{figure*}

\emph{Segment Loss. }
Next, we use the 3D consistency of 2DGS to optimize the consistency of segmentation results on the scene images. The L1 loss and SSIM loss in the original 2DGS would lead to the inability to fit object labels. To address this issue, we introduce classification loss and 3D cosine similarity loss. Specifically: 
1) for classification loss, we input the rendered identity vectors \(O\) into a linear layer\(f\) followed by a softmax operation:

\begin{equation}
F(O) = \text{softmax}(f(O))
\end{equation}

Then we use the standard cross-entropy loss \(L_{oe}\) for classification; 
2) For the 3D cosine similarity loss, we sample \(m\) 2D Gaussians, ensuring that the cosine similarity of the identity features \(F_o\) from the \(n\) nearest 2D Gaussians is closely aligned:

\begin{align*}
L_{cs} &=  \frac{1}{mn} \sum_{j=1}^{m} \sum_{i=1}^{n} \frac{{F(o_j) \cdot F(o_i)}}{{\|F(o_j)\| \|F(o_i)\|}} 
\end{align*}
Making similar identity vectors closer can improve the 3D consistency of segmentation, thus enhancing segmentation accuracy.
The total loss function is the weighted sum of the segmentation loss function and the original 2D Gaussian loss function \(L_{gs}\):
\begin{equation}
L = L_{gs} + \lambda_{oe} L_{oe} + \lambda_{cs} L_{cs}
\end{equation}
\emph{3D Consistent Segmentation. }
After training the 2DGS model to convergence, we extract the target model using its object ID. By rendering this model back into the original views, we obtain precise and 3D-consistent target masks.

\begin{figure}[h]
  \centering
  \includegraphics[width=\linewidth]{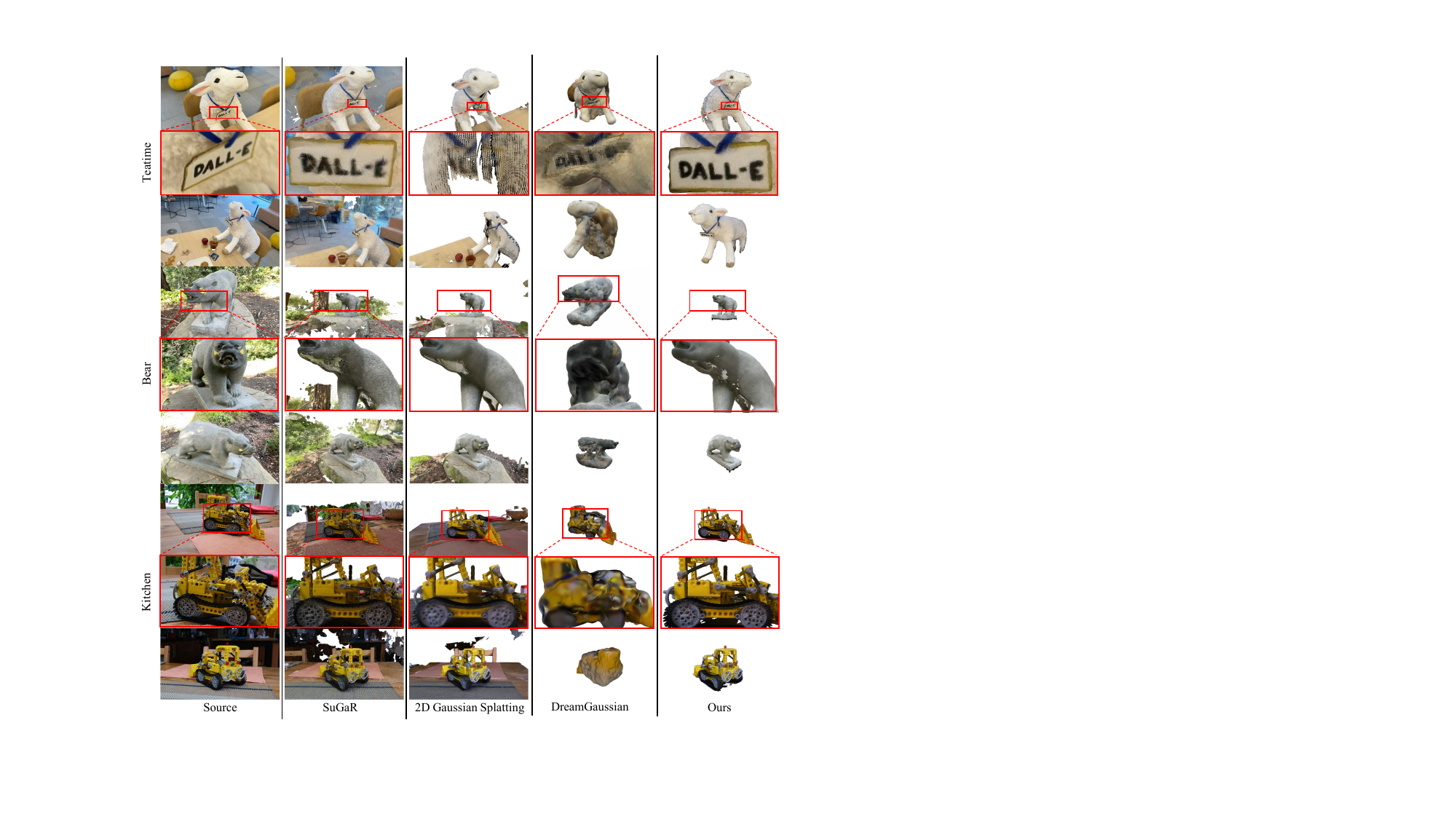}
  \caption{Comparing mesh extracting results with SuGaR, 2D Gaussian Splatting and DreamGaussain. The scenes are selected from LERF and Instruct-NeRF2NeRF datasets. Both SuGaR, 2DGS, and our approach use multi-view images as input, while DreamGaussian relies on a single-view image.}
  \label{fig:exp_1}
\end{figure}

\begin{table*}[h]
\centering
\small
\caption{Comparison of segmentation on LERF-MASK dataset~\cite{Ye2023GaussianGS}. We adopt mIoU and boundary metrics mBIoU for better segmentation quality measurement. We also compared the average memory required for training. The empirical evidence substantiates that our methodology yields superior outcomes in both quality and efficiency.}
\begin{tabular}{l|c|ccc|ccc|ccc}
\toprule
   Model & Memory & \multicolumn{3}{c|}{figurines} & \multicolumn{3}{c|}{ramen} & \multicolumn{3}{c}{teatime}  \\ 
 & GB $\downarrow$ & mIoU$\uparrow$ & mBIoU$\uparrow$ & time$\downarrow$ & mIoU$\uparrow$ & mBIoU$\uparrow$ & time$\downarrow$ & mIoU$\uparrow$ & mBIoU$\uparrow$ & time$\downarrow$\\ \midrule

SAM & & 74.83 & 73.71 &  & 57.25 & 56.99 &  & 75.95 & 74.48 &  \\ 
Gaussian Grouping (7k iter)& 21 & 86.08 & 84.08 & 7min & 70.34 & 58.03 & 7min & 75.98 & 71.91 & 7min \\
\textbf{Ours (7k iter)} & \textbf{10} & \textbf{86.21} & \textbf{84.09} & \textbf{6min} & \textbf{86.48} & \textbf{73.62} & \textbf{5min} & \textbf{78.81} & \textbf{73.32} &\textbf{6min} \\
\textbf{Ours (30k iter)} &\textbf{13} & \textbf{88.86} & \textbf{86.65} & \textbf{65min} & \textbf{88.63} & \textbf{75.93} & \textbf{51min} & \textbf{79.41} & \textbf{73.37} &\textbf{64min} \\
\bottomrule
\end{tabular}
\label{tab:seg_comp}
\end{table*}

\subsection{Target Replenishment (TR)}
As shown in the right section of Figure~\ref{pipeline}, after segmenting the target objects, we extract the target 2DGS model and obtain precise target masks in the view images. We then utilize generative diffusion priors to adaptively replenish the unseen portions and further refine the target model. In summary, we render the unseen view of the target Gaussian model, inpaint the areas that are not sufficiently reconstructed by Stable Diffusion, and then feed the inpainted images for further training.

\noindent\textbf{Stable Diffusion Personalization.} To focus the knowledge of large-scale diffusion priors on a specific target, we employ a personalization method to control Stable Diffusion. Specifically, we input the target images $\tilde{I}$, segmented from the original views, into a personalization model $\mathcal{P}$ (e.g., DreamBooth~\cite{ruiz2023dreambooth}).
\begin{equation}
\epsilon_{\theta^*} = \mathcal{P}(\tilde{I}, \epsilon_\theta)
\end{equation}
where $\epsilon_\theta$ is the UNet of original Stable Diffusion with parameter $\theta$, and $\epsilon_{\theta^*}$ is the UNet of personalized SD.

\noindent\textbf{Mask Generation \& Inpainting.}
We then use the personalized SD to generate the poorly reconstructed mask of the novel-view rendering of the target Gaussian model, as depicted in Figure~\ref{fig:method_2}. Specifically, given a novel-view image $x_0$, we encode it into latent space by the VAE encoder $\mathcal{E}$.
\begin{equation}
z_0 = \mathcal{E}(x_0)
\end{equation}
We then perturb $z_0$ by diffusion forward process with noise $\epsilon$:
\begin{equation*}
z_t = \sqrt{\overline{\alpha}_t} z_0 + \sqrt{1 - \overline{\alpha}_t} \, \epsilon, \quad \epsilon \sim \mathcal{N}(0, \mathbf{I}),
\end{equation*}
where $\mathcal{N}(0, \mathbf{I})$ is a Gaussian Distribution, $t$ denotes the timestep sampled from $[0,T]$, $z_t$ represents perturbed $z_0$ at $t$, and $\overline{\alpha}_t$ is predefined noise scheduling coefficient.

By applying the personalized UNet $\epsilon_{\theta^*}$, we can get the noise residual by:
\begin{equation*}
\scalebox{0.8}{$\triangle$} \epsilon = \epsilon_{\theta^*}(z_t, y, t) - \epsilon,
\end{equation*}
where $\scalebox{0.8}{$\triangle$} \epsilon $ is the latent noise residual, $y$ represents the embedded text prompt, which we set to empty based on experimental results.

The final inpainting mask $m$ is given by:
\begin{equation*}
m =\mathcal{B} \circ \mathcal{D}(\scalebox{0.8}{$\triangle$} \epsilon),
\end{equation*}
where $\mathcal{D}$ is the VAE decoder, and $\mathcal{B}$ is the image binarization operator.

Subsequently, the mask $m$ and the novel-view image $x_0$ are input into a standard Stable Diffusion inpainting model\footnote{huggingface.co/stabilityai/stable-diffusion-2-inpainting}, resulting in the final fixed image $\widetilde{x}_0$. 

\noindent\textbf{Mesh Extraction. }
By generating $n$ novel-view images $X = \{x_0^i\}_{i=0}^{n-1}$ and applying the inpainting process, we obtain $n$ refined images $\widetilde{X} = \{\widetilde{x}_0^i\}_{i=0}^{n-1}$. These refined images are then fed into the target 2DGS model for further training, addressing deficiencies in the first stage. 

Ultimately, following~\citeauthor{huang20242d}, to obtain meshes from the reconstructed 2D splats, we generate depth maps of the training views by projecting the splats' depth values onto the pixels. We then apply truncated signed distance fusion (TSDF) using Open3D to merge the reconstructed depth maps.

\begin{table}
\centering
\small
\caption{Quantitative results of target mesh extraction on the Tanks \& Temples dataset~\cite{knapitsch2017tanks}, represented by F1 score.
}
\begin{tabular}{l|c|c|c|c}
\toprule
   Model & {Truck} & {Ignatius} & {Caterpillar} & {Mean}  \\ 
\midrule

SuGaR & 0.0741  & 0.1392  & 0.1827  & 0.1320  \\ 
Dream Gaussian & 0.0013 & 0.0007  & 0.0031 & 0.0017   \\
2DGS & 0.1421   & 0.3423  & 0.2100  & 0.2315   \\
\textbf{Ours} & \textbf{0.1635}  & \textbf{0.4464}   & \textbf{0.2208}  &  \textbf{0.2769}  \\

\bottomrule
\end{tabular}
\label{tab:mesh_comp}
\end{table}

\section{Experiments}
\subsection{Implementation Details}

All experiments are performed and measured on a single GPU NVIDIA RTX 3090 with Ubuntu 18.04 LTS. 

\noindent\textbf{Datasets. } To evaluate the reconstruction quality, we tested our OMEGAS on scenes presented in LERF-MASK dataset ~\cite{Ye2023GaussianGS} and Mip-Nerf360 dataset~\cite{barron2022mip}, where the flowers and treehill are skipped due to the non-public access right. We also take diverse 3D scene cases from LERF~\cite{lerf2023}, Tanks\&Temples ~\cite{knapitsch2017tanks} and Instruct-NeRF2NeRF~\cite{instructnerf2023} for visual comparison. 

\noindent\textbf{Model Details. } In Target Gaussian Segmentation, we take SAM-HQ model~\cite{sam_hq} for initial segmentation The confidence threshold \(p_{ex}\) for extraction is 0.95. We use the Adam ~\cite{Kingma2014AdamAM} optimizer for both Gaussians and linear and train for max 30000 iterations with a learning rate of 0.0025 for identity vectors and 0.0005 for linear layer. For 3D regularization loss, we choose $n = 5$ and $m = 1000$. We employ Principal Component Analysis as a means to visualize the outcomes of the segmentation. 

In Target Replenishment, we adopt Stable Diffusion 2.1 with the resolution 512x512. We take DreamBooth~\cite{ruiz2023dreambooth} as our personalization model. When generating a mask, we fix the timestep $t$ to 991 and sample 10 random seeds to obtain an averaged mask result.


\begin{figure}[h]
  \centering
  \includegraphics[width=\linewidth]{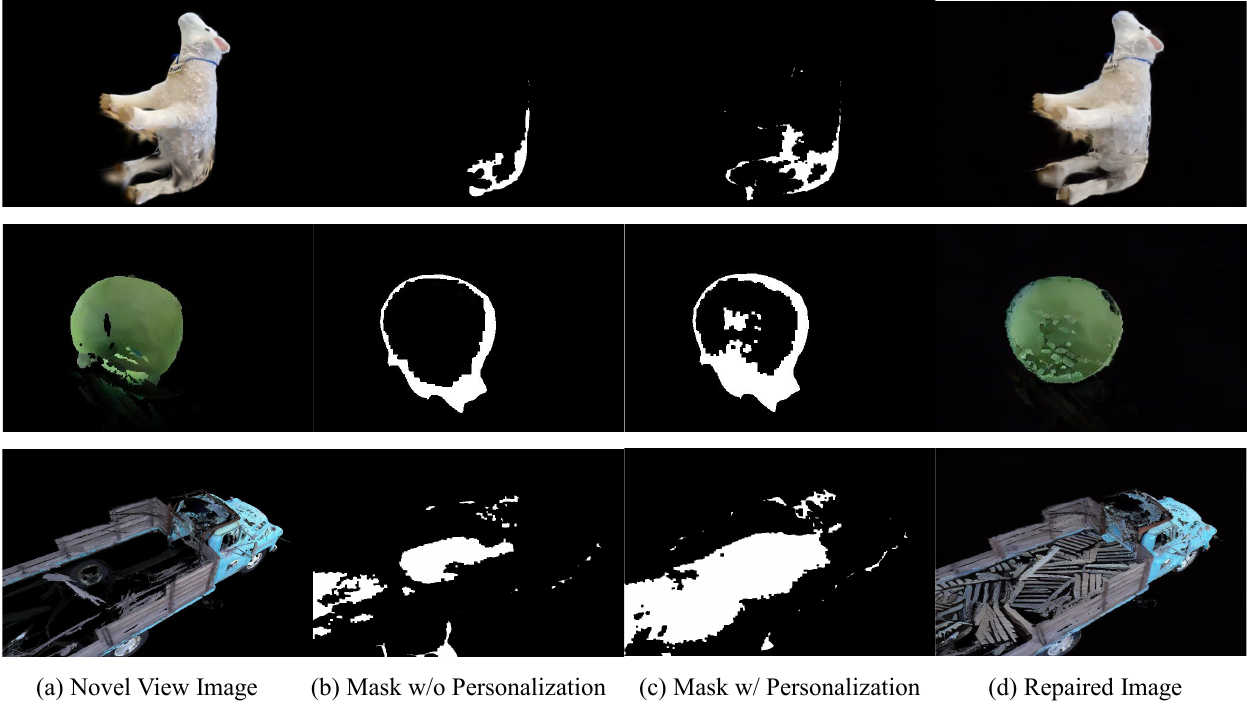}
  \caption{Effect of personalization and inpainting. It shows the mask w/ or w/o personalization of SD and the inpainting results on the Tanks\&Temples dataset and LERF dataset.}
  \label{ab:fig_inpaint}
\end{figure}

\begin{figure*}[h]
  \centering
  \includegraphics[width=\linewidth]{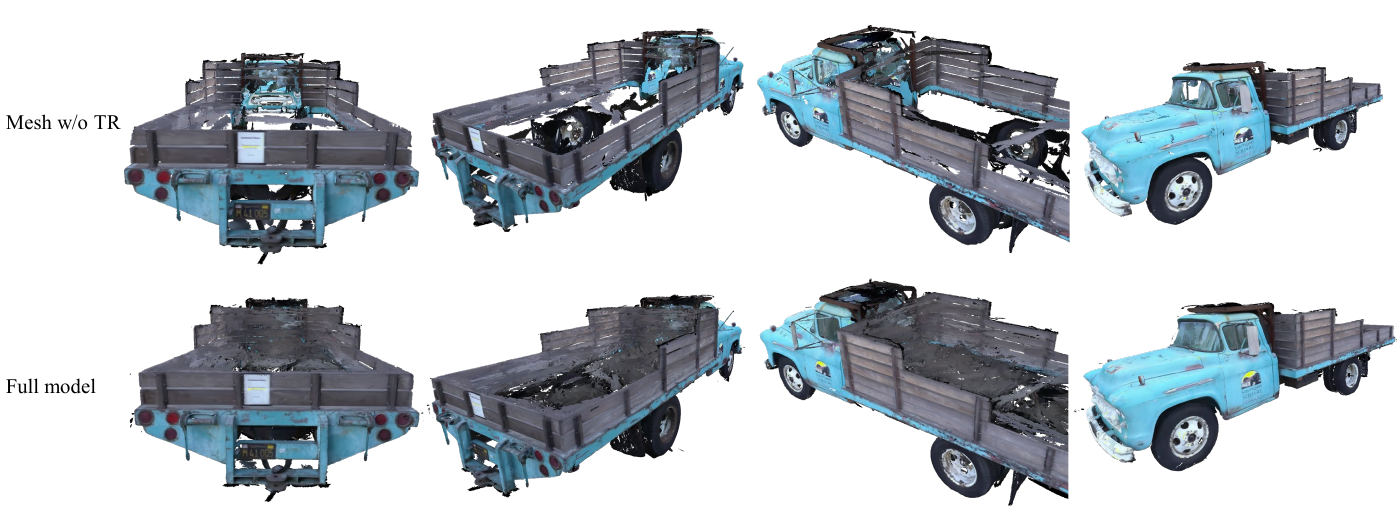}
  \caption{Effect of the Target Replenishment (TR) Module. The first line shows the mesh generated solely using Target Gaussian Segmentation. The second line displays the mesh produced by our complete model, including the Target Replenishment module.}
  \label{ab:fig_1}
\end{figure*}

\subsection{Comparative Analysis}

\noindent\textbf{Target Mesh Extraction.} To demonstrate the effectiveness of our target mesh extraction method, we provide qualitative comparisons with the scene reconstruction approach SuGaR~\cite{Guedon2023SuGaRSG} and 2DGS~\cite{huang20242d}. Since our method is the first to tackle target reconstruction in a multi-view context,  we are limited to using the single-view method DreamGaussian~\cite{tang2023dreamgaussian} as our other target reconstruction comparative baseline. The results are presented in Figure~\ref{fig:exp_1}, showcasing detailed qualitative outcomes across various open-world scenarios, including teatime, bear, and kitchen from the LERF~\cite{lerf2023} and Instruct-NeRF2NeRF~\cite{instructnerf2023} datasets. Both SuGaR, 2DGS, and our approach use multi-view images as input, while DreamGaussian relies on a single-view image.

Compared to SuGaR and 2DGS, our method achieves superior detail across all three scenes. SuGaR and 2DGS's focus on the overall scene lead to a noticeable reduction in the granularity of local details. In the bear scene, our method demonstrates its capability to reconstruct unseen portions in the input view. Meanwhile, the kitchen scene highlights the precision of our approach in segmenting and extracting complex objects. Compared to DreamGaussian, our method significantly enhances the quality of the generated meshes.

We also conduct quantitative experiments in Table \ref{tab:mesh_comp} on the Tanks \& Temples dataset~\cite{knapitsch2017tanks}. Since our method focuses on mesh extraction of target objects within a scene, we manually removed other objects from the ground truth mesh to quantify the accuracy of our approach. For methods that cannot isolate the target object, we used the complete ground truth mesh to ensure fairness. We employed the evaluation metrics provided by the Tanks\&Temples dataset~\cite{knapitsch2017tanks}, calculating precision and recall to obtain the F1 score. Our method consistently achieves superior results across all scenarios in the dataset compared to other methods.

In Figure \ref{fig:exp_2}, we further demonstrate the mesh quality and highlight our method's effectiveness in extracting small objects from complex scenes.

\noindent\textbf{Scene Segmentation. }
To demonstrate the effectiveness of our Gaussian Segmentation method, we conducted quantitative scene segmentation experiments on the LERF-MASK dataset~\cite{lerf2023}, using Gaussian Grouping~\cite{Ye2023GaussianGS}, as shown in Table \ref{tab:seg_comp}. Due to VRAM limitations, we trained Gaussian Grouping for a max of 7,000 iterations on a single NVIDIA RTX 3090 GPU. Despite these constraints, our method achieved considerable improvements under the same iteration settings, with lower memory usage and reduced training time across three scenes. When further training our method for 30,000 iterations, we observed more performance gains.

\subsection{Ablations}
In this section, we isolate the design choices and evaluate their impacts, including the effect of the Target Replenishment stage and the necessity of personalizing SD.

\noindent\textbf{Effect of Target Replenishment (TR). } We first examine the effects of the proposed Target Replenishment module, as shown in Figure~\ref{ab:fig_1}. It shows the difference between mesh results w/ and w/o TR module on an example of Tanks\&Temples dataset~\cite{knapitsch2017tanks}. Specifically, there are no images taken from above or showing the interior of the truck's carriage in the given dataset views. Without TR, a significant mesh hole would appear inside the truck, whereas our method can properly fill the mesh.

\noindent\textbf{Effect of Personalizing SD.} We now analyze the effect of personalizing SD to generate inpainting masks in Figure \ref{ab:fig_inpaint}. In the figure, (a) are images of the object rendered from a random viewpoint, (b) are masks generated by a standard Stable Diffusion model, (c) are masks generated by a personalized Stable Diffusion model, and (d) are inpainted images by mask (c). It can be observed that the mask generated by the personalized Stable Diffusion model covers a larger area and more accurately identifies the locations of the holes, resulting in a better-repaired image.

\section{Conclusion}
We present OMEGAS: Object Mesh Extraction from Large Scenes Guided by Gaussian Segmentation. OMEGAS efficiently extracts high-precision meshes of target objects from multi-view scene images and can reconstruct occluded or invisible parts of the targets. OMEGAS introduces a novel Target Gaussian Segmentation technique, which segments the target Gaussian model from multi-view images using 2D Gaussian Splatting. Additionally, OMEGAS proposes a Target Replenishment technique that leverages open-world diffusion priors to address unseen portions of the target.

\bibliography{omegas}

\begin{thebibliography}{48}
\providecommand{\natexlab}[1]{#1}

\bibitem[{Barron et~al.(2022)Barron, Mildenhall, Verbin, Srinivasan, and
  Hedman}]{barron2022mip}
Barron, J.~T.; Mildenhall, B.; Verbin, D.; Srinivasan, P.~P.; and Hedman, P.
  2022.
\newblock Mip-nerf 360: Unbounded anti-aliased neural radiance fields.
\newblock In \emph{Proceedings of the IEEE/CVF Conference on Computer Vision
  and Pattern Recognition}, 5470--5479.

\bibitem[{Caetano, Santos, and Leit{\~a}o(2020)}]{caetano2020computational}
Caetano, I.; Santos, L.; and Leit{\~a}o, A. 2020.
\newblock Computational design in architecture: Defining parametric,
  generative, and algorithmic design.
\newblock \emph{Frontiers of Architectural Research}, 9(2): 287--300.

\bibitem[{Cao et~al.(2024)Cao, Zhou, Song, and Yang}]{cao2024controllable}
Cao, P.; Zhou, F.; Song, Q.; and Yang, L. 2024.
\newblock Controllable generation with text-to-image diffusion models: A
  survey.
\newblock \emph{arXiv preprint arXiv:2403.04279}.

\bibitem[{Cheng et~al.(2023)Cheng, Oh, Price, Schwing, and
  Lee}]{cheng2023tracking}
Cheng, H.~K.; Oh, S.~W.; Price, B.; Schwing, A.; and Lee, J.-Y. 2023.
\newblock Tracking anything with decoupled video segmentation.
\newblock In \emph{Proceedings of the IEEE/CVF International Conference on
  Computer Vision}, 1316--1326.

\bibitem[{Darmon et~al.(2022)Darmon, Bascle, Devaux, Monasse, and
  Aubry}]{darmon2022improving}
Darmon, F.; Bascle, B.; Devaux, J.-C.; Monasse, P.; and Aubry, M. 2022.
\newblock Improving neural implicit surfaces geometry with patch warping.
\newblock In \emph{Proceedings of the IEEE/CVF Conference on Computer Vision
  and Pattern Recognition}, 6260--6269.

\bibitem[{Fan et~al.(2021)Fan, Dong, Zhu, Lv, Ye, and Wang}]{fan2021scf}
Fan, S.; Dong, Q.; Zhu, F.; Lv, Y.; Ye, P.; and Wang, F.-Y. 2021.
\newblock SCF-Net: Learning spatial contextual features for large-scale point
  cloud segmentation.
\newblock In \emph{Proceedings of the IEEE/CVF Conference on Computer Vision
  and Pattern Recognition}, 14504--14513.

\bibitem[{Goesele et~al.(2007)Goesele, Snavely, Curless, Hoppe, and
  Seitz}]{goesele2007multi}
Goesele, M.; Snavely, N.; Curless, B.; Hoppe, H.; and Seitz, S.~M. 2007.
\newblock Multi-view stereo for community photo collections.
\newblock In \emph{2007 IEEE 11th International Conference on Computer Vision},
  1--8. IEEE.

\bibitem[{Gu'edon and Lepetit(2023)}]{Guedon2023SuGaRSG}
Gu'edon, A.; and Lepetit, V. 2023.
\newblock SuGaR: Surface-Aligned Gaussian Splatting for Efficient 3D Mesh
  Reconstruction and High-Quality Mesh Rendering.
\newblock \emph{ArXiv}, abs/2311.12775.

\bibitem[{Haque et~al.(2023)Haque, Tancik, Efros, Holynski, and
  Kanazawa}]{instructnerf2023}
Haque, A.; Tancik, M.; Efros, A.; Holynski, A.; and Kanazawa, A. 2023.
\newblock Instruct-NeRF2NeRF: Editing 3D Scenes with Instructions.
\newblock In \emph{ICCV}.

\bibitem[{Huang et~al.(2024)Huang, Yu, Chen, Geiger, and Gao}]{huang20242d}
Huang, B.; Yu, Z.; Chen, A.; Geiger, A.; and Gao, S. 2024.
\newblock 2d gaussian splatting for geometrically accurate radiance fields.
\newblock In \emph{ACM SIGGRAPH 2024 Conference Papers}, 1--11.

\bibitem[{Ke et~al.(2023)Ke, Ye, Danelljan, Liu, Tai, Tang, and Yu}]{sam_hq}
Ke, L.; Ye, M.; Danelljan, M.; Liu, Y.; Tai, Y.-W.; Tang, C.-K.; and Yu, F.
  2023.
\newblock Segment Anything in High Quality.
\newblock In \emph{NeurIPS}.

\bibitem[{Kerbl et~al.(2023)Kerbl, Kopanas, Leimk{\"u}hler, and
  Drettakis}]{kerbl20233d}
Kerbl, B.; Kopanas, G.; Leimk{\"u}hler, T.; and Drettakis, G. 2023.
\newblock 3d gaussian splatting for real-time radiance field rendering.
\newblock \emph{ACM Transactions on Graphics}, 42(4): 1--14.

\bibitem[{Kerr et~al.(2023)Kerr, Kim, Goldberg, Kanazawa, and
  Tancik}]{lerf2023}
Kerr, J.; Kim, C.~M.; Goldberg, K.; Kanazawa, A.; and Tancik, M. 2023.
\newblock LERF: Language Embedded Radiance Fields.
\newblock In \emph{ICCV}.

\bibitem[{Kim, Seo, and Han(2022)}]{kim2022infonerf}
Kim, M.; Seo, S.; and Han, B. 2022.
\newblock Infonerf: Ray entropy minimization for few-shot neural volume
  rendering.
\newblock In \emph{Proceedings of the IEEE/CVF Conference on Computer Vision
  and Pattern Recognition}, 12912--12921.

\bibitem[{Kingma and Ba(2014)}]{Kingma2014AdamAM}
Kingma, D.~P.; and Ba, J. 2014.
\newblock Adam: A Method for Stochastic Optimization.
\newblock \emph{CoRR}, abs/1412.6980.

\bibitem[{Kirillov et~al.(2023)Kirillov, Mintun, Ravi, Mao, Rolland, Gustafson,
  Xiao, Whitehead, Berg, Lo et~al.}]{kirillov2023segment}
Kirillov, A.; Mintun, E.; Ravi, N.; Mao, H.; Rolland, C.; Gustafson, L.; Xiao,
  T.; Whitehead, S.; Berg, A.~C.; Lo, W.-Y.; et~al. 2023.
\newblock Segment anything.
\newblock In \emph{ICCV}.

\bibitem[{Knapitsch et~al.(2017)Knapitsch, Park, Zhou, and
  Koltun}]{knapitsch2017tanks}
Knapitsch, A.; Park, J.; Zhou, Q.-Y.; and Koltun, V. 2017.
\newblock Tanks and temples: Benchmarking large-scale scene reconstruction.
\newblock \emph{ACM Transactions on Graphics (ToG)}, 36(4): 1--13.

\bibitem[{Li et~al.(2023)Li, M{\"u}ller, Evans, Taylor, Unberath, Liu, and
  Lin}]{li2023neuralangelo}
Li, Z.; M{\"u}ller, T.; Evans, A.; Taylor, R.~H.; Unberath, M.; Liu, M.-Y.; and
  Lin, C.-H. 2023.
\newblock Neuralangelo: High-fidelity neural surface reconstruction.
\newblock In \emph{Proceedings of the IEEE/CVF Conference on Computer Vision
  and Pattern Recognition}, 8456--8465.

\bibitem[{Liu et~al.(2024)Liu, Xu, Jin, Chen, Varma~T, Xu, and Su}]{liu2024one}
Liu, M.; Xu, C.; Jin, H.; Chen, L.; Varma~T, M.; Xu, Z.; and Su, H. 2024.
\newblock One-2-3-45: Any single image to 3d mesh in 45 seconds without
  per-shape optimization.
\newblock \emph{Advances in Neural Information Processing Systems}, 36.

\bibitem[{Liu et~al.(2023{\natexlab{a}})Liu, Wu, Hoorick, Tokmakov, Zakharov,
  and Vondrick}]{Liu2023Zero1to3ZO}
Liu, R.; Wu, R.; Hoorick, B.~V.; Tokmakov, P.; Zakharov, S.; and Vondrick, C.
  2023{\natexlab{a}}.
\newblock Zero-1-to-3: Zero-shot One Image to 3D Object.
\newblock \emph{2023 IEEE/CVF International Conference on Computer Vision
  (ICCV)}, 9264--9275.

\bibitem[{Liu et~al.(2023{\natexlab{b}})Liu, Chen, Kao, Tai, and
  Tang}]{liu2023deceptive}
Liu, X.; Chen, J.; Kao, S.-h.; Tai, Y.-W.; and Tang, C.-K. 2023{\natexlab{b}}.
\newblock Deceptive-nerf: Enhancing nerf reconstruction using
  pseudo-observations from diffusion models.
\newblock \emph{arXiv preprint arXiv:2305.15171}.

\bibitem[{Lorensen and Cline(1987)}]{Lorensen1987MarchingCA}
Lorensen, W.~E.; and Cline, H.~E. 1987.
\newblock Marching cubes: A high resolution 3D surface construction algorithm.
\newblock \emph{Proceedings of the 14th annual conference on Computer graphics
  and interactive techniques}.

\bibitem[{Mildenhall et~al.(2020)Mildenhall, Srinivasan, Tancik, Barron,
  Ramamoorthi, and Ng}]{mildenhall2020nerf}
Mildenhall, B.; Srinivasan, P.~P.; Tancik, M.; Barron, J.~T.; Ramamoorthi, R.;
  and Ng, R. 2020.
\newblock NeRF: Representing Scenes as Neural Radiance Fields for View
  Synthesis.
\newblock In \emph{ECCV}.

\bibitem[{Mirzaei et~al.(2023)Mirzaei, Aumentado-Armstrong, Derpanis, Kelly,
  Brubaker, Gilitschenski, and Levinshtein}]{mirzaei2023spin}
Mirzaei, A.; Aumentado-Armstrong, T.; Derpanis, K.~G.; Kelly, J.; Brubaker,
  M.~A.; Gilitschenski, I.; and Levinshtein, A. 2023.
\newblock SPIn-NeRF: Multiview segmentation and perceptual inpainting with
  neural radiance fields.
\newblock In \emph{CVPR}.

\bibitem[{Niemeyer et~al.(2022)Niemeyer, Barron, Mildenhall, Sajjadi, Geiger,
  and Radwan}]{niemeyer2022regnerf}
Niemeyer, M.; Barron, J.~T.; Mildenhall, B.; Sajjadi, M.~S.; Geiger, A.; and
  Radwan, N. 2022.
\newblock Regnerf: Regularizing neural radiance fields for view synthesis from
  sparse inputs.
\newblock In \emph{Proceedings of the IEEE/CVF Conference on Computer Vision
  and Pattern Recognition}, 5480--5490.

\bibitem[{Oechsle, Peng, and Geiger(2021)}]{oechsle2021unisurf}
Oechsle, M.; Peng, S.; and Geiger, A. 2021.
\newblock Unisurf: Unifying neural implicit surfaces and radiance fields for
  multi-view reconstruction.
\newblock In \emph{Proceedings of the IEEE/CVF International Conference on
  Computer Vision}, 5589--5599.

\bibitem[{Poole et~al.(2022)Poole, Jain, Barron, and
  Mildenhall}]{poole2022dreamfusion}
Poole, B.; Jain, A.; Barron, J.~T.; and Mildenhall, B. 2022.
\newblock Dreamfusion: Text-to-3d using 2d diffusion.
\newblock \emph{arXiv preprint arXiv:2209.14988}.

\bibitem[{Rombach et~al.(2022)Rombach, Blattmann, Lorenz, Esser, and
  Ommer}]{rombach2022high}
Rombach, R.; Blattmann, A.; Lorenz, D.; Esser, P.; and Ommer, B. 2022.
\newblock High-resolution image synthesis with latent diffusion models.
\newblock In \emph{Proceedings of the IEEE/CVF conference on computer vision
  and pattern recognition}, 10684--10695.

\bibitem[{Ruiz et~al.(2023)Ruiz, Li, Jampani, Pritch, Rubinstein, and
  Aberman}]{ruiz2023dreambooth}
Ruiz, N.; Li, Y.; Jampani, V.; Pritch, Y.; Rubinstein, M.; and Aberman, K.
  2023.
\newblock Dreambooth: Fine tuning text-to-image diffusion models for
  subject-driven generation.
\newblock In \emph{Proceedings of the IEEE/CVF conference on computer vision
  and pattern recognition}, 22500--22510.

\bibitem[{Seo, Chang, and Kwak(2023)}]{seo2023flipnerf}
Seo, S.; Chang, Y.; and Kwak, N. 2023.
\newblock Flipnerf: Flipped reflection rays for few-shot novel view synthesis.
\newblock In \emph{Proceedings of the IEEE/CVF International Conference on
  Computer Vision}, 22883--22893.

\bibitem[{Siciliano, Khatib, and Kr{\"o}ger(2008)}]{siciliano2008springer}
Siciliano, B.; Khatib, O.; and Kr{\"o}ger, T. 2008.
\newblock \emph{Springer handbook of robotics}, volume 200.
\newblock Springer.

\bibitem[{Snavely, Seitz, and Szeliski(2006)}]{snavely2006photo}
Snavely, N.; Seitz, S.~M.; and Szeliski, R. 2006.
\newblock Photo tourism: exploring photo collections in 3D.
\newblock In \emph{ACM siggraph 2006 papers}, 835--846. Association for
  Computing Machinery.

\bibitem[{Song et~al.(2023)Song, Park, An, Cho, Kwak, Cho, and
  Kim}]{song2023darf}
Song, J.; Park, S.; An, H.; Cho, S.; Kwak, M.-S.; Cho, S.; and Kim, S. 2023.
\newblock D{\"a}RF: boosting radiance fields from sparse inputs with monocular
  depth adaptation.
\newblock In \emph{Proceedings of the 37th International Conference on Neural
  Information Processing Systems}, 68458--68470.

\bibitem[{Tang et~al.(2023)Tang, Ren, Zhou, Liu, and
  Zeng}]{tang2023dreamgaussian}
Tang, J.; Ren, J.; Zhou, H.; Liu, Z.; and Zeng, G. 2023.
\newblock Dreamgaussian: Generative gaussian splatting for efficient 3d content
  creation.
\newblock \emph{arXiv preprint arXiv:2309.16653}.

\bibitem[{Wang et~al.(2023)Wang, Chen, Loy, and Liu}]{wang2023sparsenerf}
Wang, G.; Chen, Z.; Loy, C.~C.; and Liu, Z. 2023.
\newblock Sparsenerf: Distilling depth ranking for few-shot novel view
  synthesis.
\newblock In \emph{Proceedings of the IEEE/CVF International Conference on
  Computer Vision}, 9065--9076.

\bibitem[{Wang et~al.(2021)Wang, Liu, Liu, Theobalt, Komura, and
  Wang}]{wang2021neus}
Wang, P.; Liu, L.; Liu, Y.; Theobalt, C.; Komura, T.; and Wang, W. 2021.
\newblock Neus: Learning neural implicit surfaces by volume rendering for
  multi-view reconstruction.
\newblock \emph{arXiv preprint arXiv:2106.10689}.

\bibitem[{Wang et~al.(2024)Wang, Lu, Wang, Bao, Li, Su, and
  Zhu}]{wang2024prolificdreamer}
Wang, Z.; Lu, C.; Wang, Y.; Bao, F.; Li, C.; Su, H.; and Zhu, J. 2024.
\newblock Prolificdreamer: High-fidelity and diverse text-to-3d generation with
  variational score distillation.
\newblock \emph{Advances in Neural Information Processing Systems}, 36.

\bibitem[{Wu et~al.(2023)Wu, Yi, Fang, Xie, Zhang, Wei, Liu, Tian, and
  Wang}]{wu20234d}
Wu, G.; Yi, T.; Fang, J.; Xie, L.; Zhang, X.; Wei, W.; Liu, W.; Tian, Q.; and
  Wang, X. 2023.
\newblock 4d gaussian splatting for real-time dynamic scene rendering.
\newblock \emph{arXiv preprint arXiv:2310.08528}.

\bibitem[{Wu et~al.(2024)Wu, Mildenhall, Henzler, Park, Gao, Watson,
  Srinivasan, Verbin, Barron, Poole et~al.}]{wu2024reconfusion}
Wu, R.; Mildenhall, B.; Henzler, P.; Park, K.; Gao, R.; Watson, D.; Srinivasan,
  P.~P.; Verbin, D.; Barron, J.~T.; Poole, B.; et~al. 2024.
\newblock Reconfusion: 3d reconstruction with diffusion priors.
\newblock In \emph{Proceedings of the IEEE/CVF Conference on Computer Vision
  and Pattern Recognition}, 21551--21561.

\bibitem[{Wynn and Turmukhambetov(2023)}]{wynn2023diffusionerf}
Wynn, J.; and Turmukhambetov, D. 2023.
\newblock Diffusionerf: Regularizing neural radiance fields with denoising
  diffusion models.
\newblock In \emph{Proceedings of the IEEE/CVF Conference on Computer Vision
  and Pattern Recognition}, 4180--4189.

\bibitem[{Xie et~al.(2023)Xie, Xu, Rakotosaona, Rim, Tombari, Keutzer,
  Tomizuka, and Zhan}]{xie2023sparsefusion}
Xie, Y.; Xu, C.; Rakotosaona, M.-J.; Rim, P.; Tombari, F.; Keutzer, K.;
  Tomizuka, M.; and Zhan, W. 2023.
\newblock Sparsefusion: Fusing multi-modal sparse representations for
  multi-sensor 3d object detection.
\newblock In \emph{Proceedings of the IEEE/CVF International Conference on
  Computer Vision}, 17591--17602.

\bibitem[{Xiong et~al.(2021)Xiong, Hsiang, He, Zhan, and
  Wu}]{xiong2021augmented}
Xiong, J.; Hsiang, E.-L.; He, Z.; Zhan, T.; and Wu, S.-T. 2021.
\newblock Augmented reality and virtual reality displays: emerging technologies
  and future perspectives.
\newblock \emph{Light: Science \& Applications}, 10(1): 1--30.

\bibitem[{Xu et~al.(2020)Xu, Wu, Wang, Zhan, Vajda, Keutzer, and
  Tomizuka}]{xu2020squeezesegv3}
Xu, C.; Wu, B.; Wang, Z.; Zhan, W.; Vajda, P.; Keutzer, K.; and Tomizuka, M.
  2020.
\newblock Squeezesegv3: Spatially-adaptive convolution for efficient
  point-cloud segmentation.
\newblock In \emph{Computer Vision--ECCV 2020: 16th European Conference,
  Glasgow, UK, August 23--28, 2020, Proceedings, Part XXVIII 16}, 1--19.
  Springer.

\bibitem[{Yang et~al.(2022)Yang, Bao, Zeng, Bao, Zhang, Cui, and
  Zhang}]{yang2022neumesh}
Yang, B.; Bao, C.; Zeng, J.; Bao, H.; Zhang, Y.; Cui, Z.; and Zhang, G. 2022.
\newblock Neumesh: Learning disentangled neural mesh-based implicit field for
  geometry and texture editing.
\newblock In \emph{European Conference on Computer Vision}, 597--614. Springer.

\bibitem[{Yang et~al.(2024)Yang, Li, Fang, Liang, Xie, Zhang, Shen, and
  Tian}]{yang2024gaussianobject}
Yang, C.; Li, S.; Fang, J.; Liang, R.; Xie, L.; Zhang, X.; Shen, W.; and Tian,
  Q. 2024.
\newblock Gaussianobject: Just taking four images to get a high-quality 3d
  object with gaussian splatting.
\newblock \emph{arXiv preprint arXiv:2402.10259}.

\bibitem[{Yang et~al.(2023)Yang, Gao, Zhou, Jiao, Zhang, and
  Jin}]{yang2023deformable}
Yang, Z.; Gao, X.; Zhou, W.; Jiao, S.; Zhang, Y.; and Jin, X. 2023.
\newblock Deformable 3d gaussians for high-fidelity monocular dynamic scene
  reconstruction.
\newblock \emph{arXiv preprint arXiv:2309.13101}.

\bibitem[{Ye et~al.(2023)Ye, Danelljan, Yu, and Ke}]{Ye2023GaussianGS}
Ye, M.; Danelljan, M.; Yu, F.; and Ke, L. 2023.
\newblock Gaussian Grouping: Segment and Edit Anything in 3D Scenes.
\newblock \emph{ArXiv}, abs/2312.00732.

\bibitem[{Yu et~al.(2021)Yu, Ye, Tancik, and Kanazawa}]{yu2021pixelnerf}
Yu, A.; Ye, V.; Tancik, M.; and Kanazawa, A. 2021.
\newblock pixelnerf: Neural radiance fields from one or few images.
\newblock In \emph{Proceedings of the IEEE/CVF Conference on Computer Vision
  and Pattern Recognition}, 4578--4587.

\end{thebibliography}

\end{document}